\documentclass{article}

\usepackage{microtype}
\usepackage{graphicx}
\usepackage{subfigure}
\usepackage{booktabs} 
\usepackage[usenames,dvipsnames]{xcolor}

\usepackage{hyperref}

\usepackage[accepted]{icml2021}

\usepackage{xspace}
\usepackage{paralist}
\usepackage{multirow}
\usepackage{amsmath}
\usepackage{amsfonts}
\usepackage{amssymb}
\usepackage{mathrsfs}
\usepackage{mdwlist}

\usepackage{hyperref}       
\usepackage{url}            
\usepackage{svg}
\usepackage{booktabs}       
\usepackage{bm}
\usepackage{wrapfig} 
\usepackage{amsfonts} 
\usepackage{natbib}
\usepackage{amsthm}
\usepackage{amsmath}
\usepackage{amssymb}
\usepackage{multirow}
\usepackage{microtype}      
\usepackage{subfigure}
\usepackage{longtable}
\usepackage{algorithm}
\usepackage{algorithmic}
\usepackage{wrapfig}

\usepackage{todonotes}

\icmltitlerunning{KNAS: Green Neural Architecture Search}

\begin{document}

\twocolumn[
\icmltitle{KNAS: Green Neural Architecture Search}



\icmlsetsymbol{equal}{*}

\begin{icmlauthorlist}
\icmlauthor{Jingjing Xu}{goo,equal}
\icmlauthor{Liang Zhao}{bc}
\icmlauthor{Junyang Lin}{ed}
\icmlauthor{Rundong Gao}{goo}
\icmlauthor{Xu Sun}{goo,bc}
\icmlauthor{Hongxia Yang}{ed}
\end{icmlauthorlist}

\icmlaffiliation{goo}{MOE Key Lab of Computational Linguistics, School of EECS, Peking University}
\icmlaffiliation{ed}{Alibaba Group}
\icmlaffiliation{bc}{Center for Data Science, Peking University}
\icmlcorrespondingauthor{Jingjing Xu}{jingjingxu@pku.edu.cn}
\icmlcorrespondingauthor{Xu Sun}{xusun@pku.edu.cn}

\icmlkeywords{Machine Learning, ICML}

\vskip 0.3in
]



\printAffiliationsAndNotice{\icmlEqualwork} 

\begin{abstract}

Many existing neural architecture search (NAS) solutions rely on downstream training for architecture evaluation, which takes enormous computations. Considering that these computations bring a large carbon footprint, this paper aims to explore a green (namely environmental-friendly) NAS solution that evaluates architectures without training. Intuitively, gradients, induced by the architecture itself, directly decide the convergence and generalization results. It motivates us to propose the gradient kernel hypothesis: Gradients can be used as a coarse-grained proxy of downstream training to evaluate random-initialized networks. To support the hypothesis, we conduct a theoretical analysis and find a practical gradient kernel that has good correlations with training loss and validation performance. According to this hypothesis, we propose a new kernel based architecture search approach KNAS.  Experiments show that KNAS achieves competitive results with orders of magnitude faster than ``train-then-test'' paradigms on image classification tasks. Furthermore, the extremely low search cost enables its wide applications. The searched network also outperforms strong baseline RoBERTA-large on two text classification tasks. Codes are available at \url{https://github.com/Jingjing-NLP/KNAS}.

\end{abstract}

\section{Introduction}
\label{sec:intro}

Neural architecture search (NAS) is a field to automatically explore the optimal architecture~\citep{DBLP:conf/iclr/ZophL17,DBLP:conf/iclr/BakerGNR17}. The search procedure can be divided into three components: search space, optimization approaches, architecture evaluation. Search space defines all network candidates that can be examined to produce the final
architecture. The optimization method dictates how to explore the search space. Architecture evaluation is responsible for helping the optimization method to evaluate the quality of architectures. Recent NAS approaches have been able to generate state-of-the-art models on downstream tasks~\cite{zoph2018learning,tan2019efficientnet}. 

However, despite good performance, NAS usually requires huge computations to find the optimal architecture, most of which are used for architecture evaluation. For example, \citet{DBLP:conf/iclr/ZophL17} use 800
GPUs for 28 days resulting in 22,400 GPU-hours. ~\citet{DBLP:conf/acl/StrubellGM19} find that a single neural architecture search solution can emit as much carbon as five cars in their lifetimes.  The major recent advance of NAS is to reduce the search costs. One research line focuses on search methods without architecture evaluation, like DART~\citep{DBLP:conf/iclr/LiuSY19,chu2020fair}.  While being simple, previous studies have shown these approaches suffer from well-known performance collapse due to an inevitable aggregation of skip connections~\citep{DBLP:conf/icml/YingKCR0H19,DBLP:conf/eccv/ChuZZL20}. Another line focuses on reducing architecture evaluation costs by using techniques like early-stopping~\citep{liang2019darts}, weight-sharing~\citep{pham2018efficient}, performance predicting~\cite{DBLP:conf/eccv/LiuZNSHLFYHM18}, and so on.  Compared to the original full-training solution, these techniques achieve promising speedup. However, if we consider a complicated search space with plenty of networks, they still require many computations.

\begin{figure}[t]
\centering
\begin{minipage}{.23\textwidth}
\centering
\includegraphics[width=\textwidth]{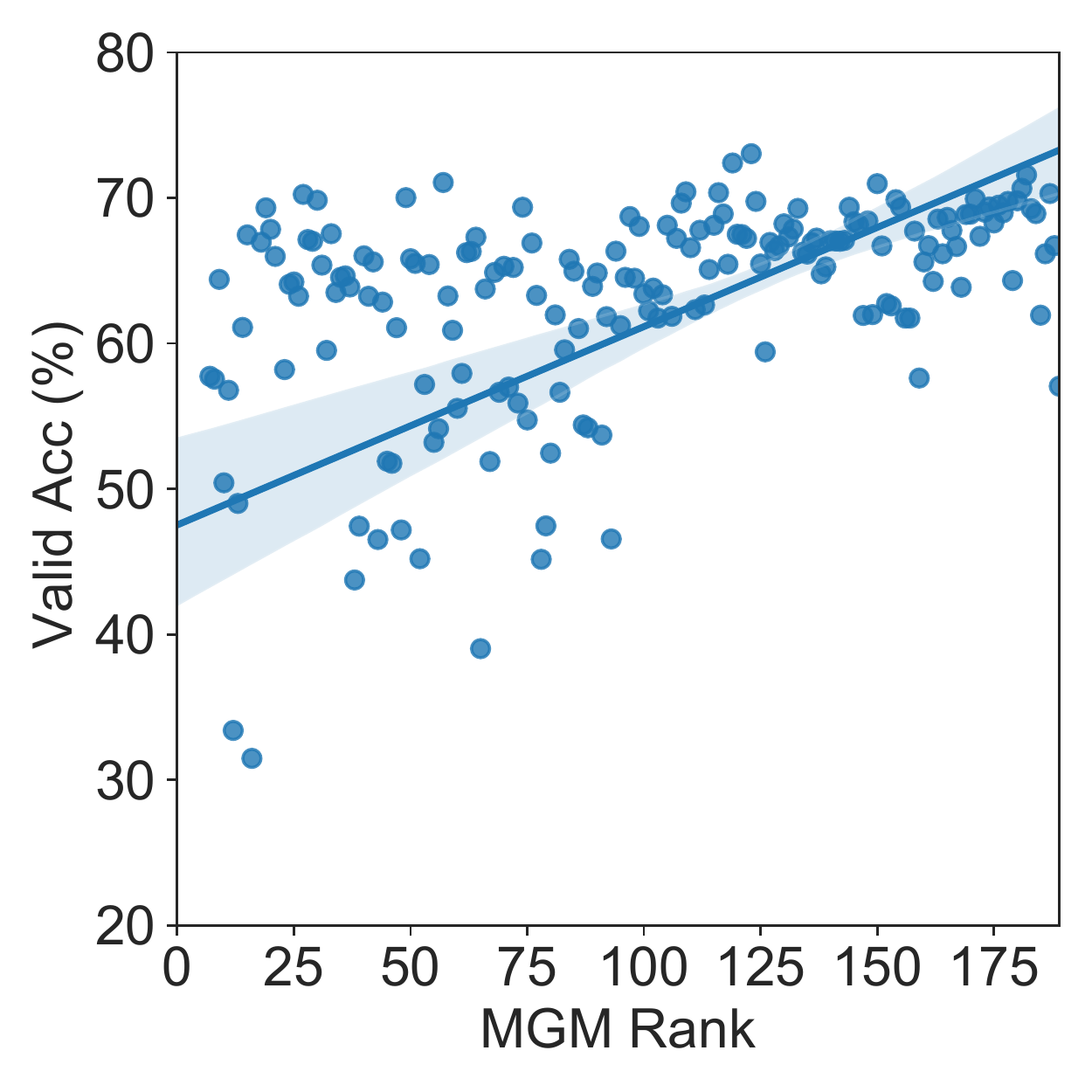}
\label{fig:prob1_6_2}
\end{minipage}
 \hfill
\begin{minipage}{0.23\textwidth}
\centering
\includegraphics[width=\linewidth]{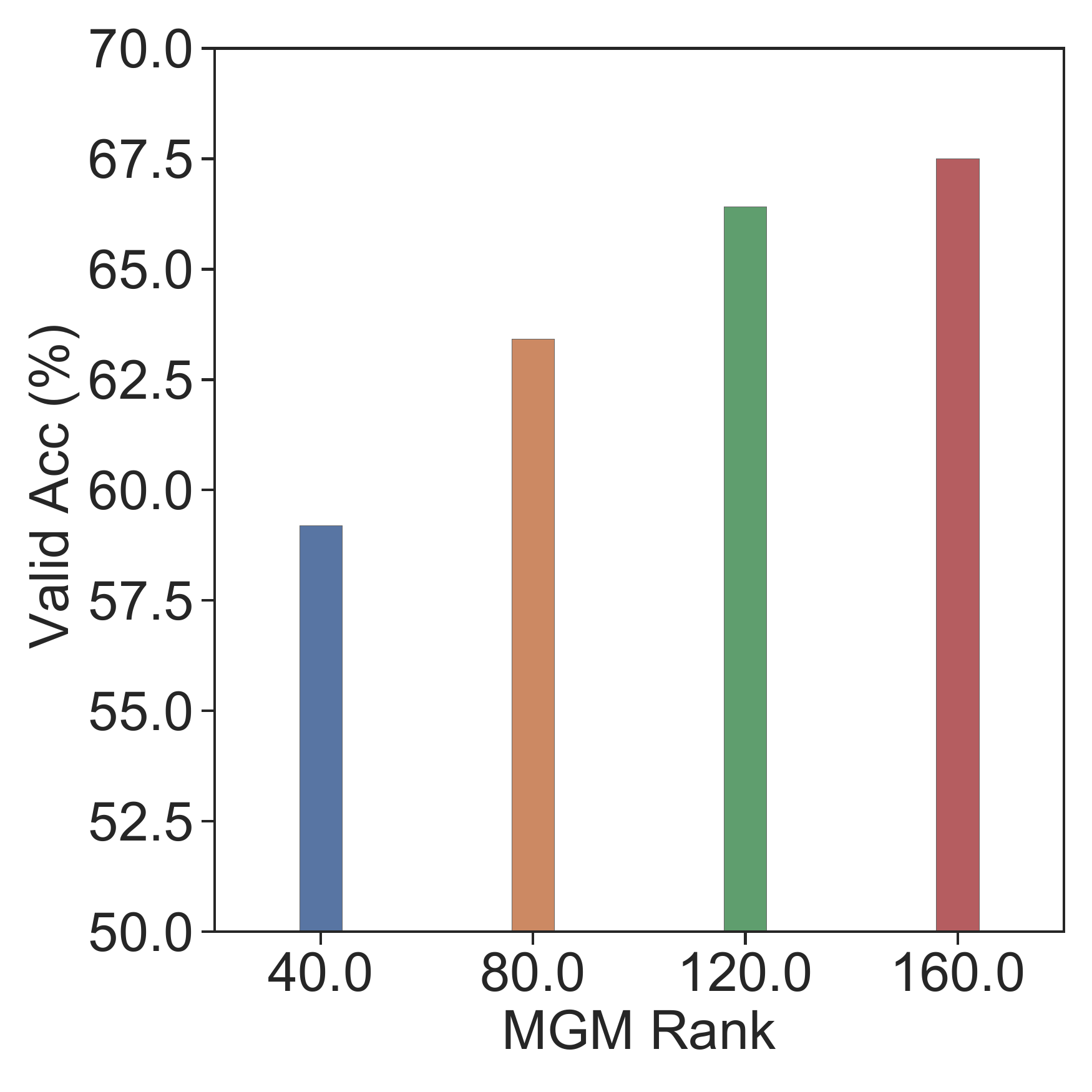}
\label{fig:prob1_6_1}
\end{minipage}
\caption{An illustration of the relation between MGM and accuracy. We sample 200 architectures and rank them based on MGM scores. Smaller rank represents smaller MGM. Y-axis lists the validation accuracy. The right figure classifies architectures into four groups based on their MGM ranks. Each bar represents the average accuracy. The Spearman coefficient is 0.56  ($P \ll 0.01 $), indicating good positive correlations.  }
\label{fig:illu}
\end{figure}

In this work, we aim to explore a more challenging question: Can we evaluate architectures without training? 
Before answering this question, it is essential to figure out how architecture affects optimization. 
It is widely accepted that gradients, induced by neural networks, play a crucial role in optimization~\citep{DBLP:journals/tnn/BengioSF94,DBLP:journals/neco/HochreiterS97,DBLP:conf/icml/PascanuMB13}. Motivated by this common belief, we propose a bold hypothesis: gradients can be used as a coarse-grained proxy of downstream training to evaluate randomly-initialized architectures. To support the hypothesis, we conduct a comprehensive theoretical analysis and identify a key gradient feature, the Gram matrix of gradients (GM). For any neural networks, the F-norm of GM decides the upper bound of convergence rate. Higher F-norm is expected for a higher convergence rate.  Intuitively, GM can be regarded as a health index of gradients. Each element in GM is the dot product between any two gradient vectors. Vanishing gradients and ataxic gradients both get lower GM scores. Following the theoretical results, we propose to leverage GM  to evaluate randomly-initialized networks. To be specific, the mean of GM, short for MGM,  is adopted in implementation. We conduct experiments on CIFAR100 and calculate the Spearman correlation scores between MGM and key optimization metrics. Experiments show that MGM has good correlations with negative training loss and validation accuracy, with 0.53  ($P \ll 0.01 $) and 0.56  ($P \ll 0.01 $) coefficients, respectively\footnote{\url{https://www.statstutor.ac.uk/resources/uploaded/spearmans.pdf}}. Figure~\ref{fig:illu} illustrates the relation between MGM and validation accuracy. These results are strong evidence to support the hypothesis, demonstrating that MGM has good potential to be used as a coarse-grained architecture evaluation. 

According to our hypothesis, we propose a gradient kernel based NAS solution, short for KNAS. We first select top-$k$ architectures with the largest MGM as candidates, which are then trained to choose the best one. In practice, $k$ is usually set to be a very small value. Experiments show that the new approach achieves large speedups with competitive accuracies on NAS-Bench-201~\citep{DBLP:conf/iclr/Dong020}, a NAS benchmark dataset. Furthermore, the low search cost allows us to apply KNAS to diverse tasks. We find that the structures searched by KNAS outperform strong baseline RoBERTA-large on two text classification tasks.  

The main contributions are summarized as follows:
\begin{itemize}

\item We propose a gradient kernel hypothesis to explore training-free architecture evaluation approaches. 

\item We find a practical gradient feature MGM to support the hypothesis.

\item Based on the hypothesis, we propose a green NAS solution KNAS that can find well-performing architectures with orders of magnitude faster than ``train-then-test'' NAS solutions. 
\end{itemize}

\section{Related Work}
\label{sec:related}

\paragraph{Network architecture search}
Neural architecture search aims to replace expert-designed networks with learned architectures~\citep{DBLP:conf/nips/ChangZGMXP19,DBLP:journals/corr/abs-1912-00195,DBLP:journals/corr/abs-2001-01233,DBLP:journals/corr/abs-2004-08546,DBLP:conf/iclr/FangSPZL0W20,DBLP:journals/corr/abs-2003-12238, DBLP:journals/corr/abs-2003-11236, DBLP:journals/corr/abs-2005-07669}. 
The first NAS research line mainly focuses on random search. 
The key idea is to randomly evaluate various architectures and select the best one based on their validation results. 
However, despite promising results, many of them require thousands of GPU days to achieve desired results.  To address this problem, \citet{DBLP:conf/iclr/ZophL17} propose a reinforcement learning based search policy that introduces an architecture generator with validation accuracy as a reward.  

Another research line is based on evolution approaches. ~\citet{DBLP:conf/aaai/RealAHL19} propose a two-stage search policy. The first stage selects several well-performing parent architectures. The second stage applies mutation on these parent architectures to select the best one.  Following this work, ~\citet{DBLP:conf/icml/SoLL19} apply the evolution search on Transformer networks and achieve new state-of-the-art results on machine translation and language modeling tasks. Although these approaches can reduce exploration costs, the dependence on downstream training still leads to huge computation costs.

To fully get rid of the dependence on validation accuracy, several studies~\citep{DBLP:conf/emnlp/JiangHXZZ19,DBLP:conf/iclr/LiuSY19,DBLP:conf/cvpr/DongY19,DBLP:conf/iclr/ZelaESMBH20,DBLP:journals/corr/abs-2005-03566} re-formulate
the task in a differentiable manner and allow efficient search using gradient descent. Furthermore, 
Unlike these studies, we propose a lightweight NAS approach, which largely reduces evaluation costs. 

There are three corresponding studies similar to our approach~\citep{DBLP:journals/corr/abs-2006-04647,DBLP:journals/corr/abs-2102-11535,DBLP:journals/corr/abs-2101-08134}. 
~\citet{DBLP:journals/corr/abs-2006-04647} evaluate randomly-initialized architectures based on the output of rectified linear units.  In contrast, KNAS uses gradient kernels to evaluate architectures. Moreover, KNAS does not require any architecture constraints.  ~\citet{DBLP:journals/corr/abs-2102-11535} propose to rank randomly-initialized architectures by analyzing the spectrum of the neural tangent kernel and the number of linear regions in the input space. Different from this approach, KNAS uses the Gram matrix of gradients to rank architectures. ~\citet{DBLP:journals/corr/abs-2101-08134} evaluates several conventional reduced-training proxies for ranking randomly-initialized models.  

\paragraph{Understanding and improving optimization}
Understanding and improving the optimization of deep networks has long been a hot research topic. ~\citet{DBLP:journals/tnn/BengioSF94} find the vanishing and exploding gradient problem in neural network training. A lot of advanced optimization approaches have been proposed in recent years, which can be classified into four research lines.  
The first research line focuses on initialization~\citep{DBLP:conf/icml/SutskeverMDH13,DBLP:journals/corr/MishkinM15,DBLP:conf/nips/HaninR18}. \citet{DBLP:journals/jmlr/GlorotB10} propose to control the variance of parameters via appropriate initialization. Following this work, several widely-used initialization approaches have been proposed, including Kaiming initialization~\citep{DBLP:conf/iccv/HeZRS15}, Xaiver initinalization~\citep{DBLP:journals/jmlr/GlorotB10}, and Fixup initialization~\citep{DBLP:conf/iclr/ZhangDM19}. The second research line focuses on normalization~\citep{ioffe2015batch,lei2016layer,DBLP:journals/corr/UlyanovVL16,DBLP:conf/eccv/WuH18,DBLP:journals/corr/abs-1910-05895}. These approaches aim to avoid the vanishing gradient by controlling the distribution of intermediate layers.
The third is mainly based on activation functions to make the derivatives of activation less saturated to avoid the vanishing problem, including GELU activation~\citep{hendrycks2016gelu}, SELU activation~\citep{DBLP:conf/nips/KlambauerUMH17}, and so on. The motivation behind these approaches is to avoid unsteady derivatives of the activation concerning the inputs. 
The fourth focuses on gradient clipping~\citep{DBLP:conf/icml/PascanuMB13}. 

\paragraph{Gradient Kernel}
Our work is also related to gradient kernels~\citep{DBLP:journals/corr/abs-1710-03667,DBLP:conf/nips/JacotHG18}.  NTK~\citep{DBLP:conf/nips/JacotHG18,DBLP:conf/iclr/DuZPS19} is a popular gradient kernel, defined as the Gram matrix of gradients. It is proposed to analyze the model's convergence and generalization. Following these studies, many researchers are devoted to understand current networks from the perspective of NTK~\citep{DBLP:conf/nips/LeeXSBNSP19, DBLP:journals/corr/abs-1903-08560, DBLP:conf/nips/Allen-ZhuLL19,DBLP:conf/icml/AroraDHLW19}. ~\citet{DBLP:conf/iclr/DuZPS19} use the Gram matrix of gradients to prove that for an $m$ hidden node shallow neural network with ReLU activation, as long as $m$ is large enough, randomly initialized gradient descent converges to a globally optimal solution at a linear convergence rate for the quadratic loss function. Following this work, ~\citet{DBLP:conf/icml/DuLL0Z19} further expand this finding and prove that gradient descent achieves zero training loss in polynomial time for a deep over-parameterized neural network with residual connections. 

\section{Gradient Kernel Hypothesis}

In this section, we introduce a gradient kernel hypothesis for training-free architecture evaluation. It is widely believed that 
gradients, induced by neural networks, are the most direct factor to decide convergence and generalization results~\citep{DBLP:journals/siamjo/YuanLY16}. It motivates us to explore whether gradients can be used as an alternative replacement of downstream training to evaluate architectures. 

We consider a simple multi-layer fully-connected neural network with $L$ layers. Assume that the output of $h$-th layer is:
\begin{equation}
\bm y^{(h)} = \sigma(\bm w^{(h)} \bm y^{(h-1)}),
\end{equation}
where $\sigma$ is the activation function and $ \bm w^{(h)}$ is the weight matrix.  The gradient for the $h$-th layer with respect to the weight matrix $ \bm w^{(h)}$ is: 
\begin{equation}
\begin{split}
\frac{\partial \mathcal{L}}{\partial \bm w^{(h)}}  &= \langle \frac{\partial \mathcal{L}}{\partial \bm y^{(L)}},   \bm y^{(h-1)} \frac{\partial{\bm y^{(L)}}}{\partial \bm y^{(h)}}\bm J^{(h)}  \rangle  \\
&=\langle \frac{\partial \mathcal{L}}{\partial \bm y^{(L)}},    \bm y^{(h-1)} ( \prod_{k=h+1}^{L}\bm J^{(k)}  \bm w^{(k)} ) \bm J^{(h)}  \rangle,
\label{eq:par}
\end{split}
\end{equation}
where $\mathcal{L}$ is the loss function and $\frac{\partial \mathcal{L}}{\partial \bm y^{(L)}}$ is the derivative of the output of the last layer\footnote{Here we adopt numerator layout notation.}. $\bm y^{(h-1)}$ is the output of $h-1$ layer. $\bm J^{(k)}$ is the diagonal matrix where the main diagonal is the list of the derivatives of the activation with respect to the inputs in the $k$-th layer:
\begin{equation}
    \bm J^{(k)} = \textbf{diag}(\sigma{'}( \bm w_1^{(k)}\bm y^{(k-1)}),  \cdots, \sigma{'}( \bm w_d^{(k)} \bm y^{(k-1)})),
\end{equation}
where $d$ is the dimension of the activation and $\bm w_d^{(k)}$ is the $d$-th row in matrix $\bm w^{(k)}$ in the $k$ layer.

From this equation, we can see that gradients depend on the design of networks, including network depth, activation function, initialization, and so on. Therefore, different architectures will result in totally different gradients. Motivated by this, we propose a gradient kernel hypothesis:
\newtheorem{prop}{Proposition}[section]
\begin{prop}[Gradient Kernel Hypothesis]
Suppose $\mathcal{G}$ is a set of all gradient features. There exists a gradient feature $g \in \mathcal{G}$ that can be used as a coarse-grained proxy of downstream training to evaluate randomly-initialized architectures.
\end{prop}


\section{Identify Key Gradient Kernels}
\label{sec:theory}
To support the hypothesis, we follow the idea of NTK~\citep{DBLP:conf/nips/JacotHG18} and show an theoretical understanding about the relationship between architectures and convergence results. Theoretical results show that the Gram matrix of gradients, short for GM, decides the convergence results. It is a good signal showing that GM is likely to be a good proxy of downstream performance to evaluate the quality of architectures.

Formally, we consider a neural network with $L$ layers. On the $h$-th layer, the output of $\bm y^{(h+1)}$ is computed as:
\begin{equation}
    \bm y^{(h+1)} = \sigma( \bm w^{(h)}\bm y^{(h)}),
\end{equation}
where $\bm y_0=\bm x$ and $\bm x \in \mathcal{R}^{d}$.  $\bm w^{(h)} \in \mathcal{R}^{d \times d}$ is a trainable parameter matrix. The last layer is a sum layer responsible for generating the final label $y^{(L)}$. We focus on the empirical risk minimization problem with a quadratic loss. 
Given a training dataset ${(\bm x, \bm y)^n_{i=1}}$, we train the model by minimizing MSE loss:
\begin{equation}
    \mathcal{L}(\bm w) = \frac{1}{2}\| \bm y^{(L)} -  \bm y^{*}\|_2^2,
\end{equation}
where $\bm y^{*}= \{y_{1}^{*},\cdots,y_{n}^{*}\}$ is the gold label vector. $\bm y = \{y^{(L)}_{1},\cdots,y^{(L)}_{n}\} $ is the prediction label vector. $n$ is the size of training data.
We apply gradient descent to optimize weights:
\begin{equation}
   \bm w(t+1) = \bm w(t) - \mu  \frac{\partial \mathcal{L}(\bm w(t))}{\partial \bm w(t)},
\end{equation}
where $t$ represent the $t$-th iteration,  $\mu > 0$ is the step size, $\bm w(t)$ is the parameter vector at the $t$-th iteration. The gradient vector is:
\begin{equation}
 \frac{\partial \mathcal{L}(\bm w(t),i)}{\partial( \bm w(t))} =(y^{(L)}_{i} - y_{i}^{*})\frac{\partial{ y^{(L)}_{i}}}{\partial \bm w(t)},
\end{equation}
where $\mathcal{L}(\bm w(t),i)$ is the loss function of example $i$.
We consider gradient descent with infinitesimal step size following~\citet{DBLP:conf/icml/AroraCH18}.  Formally, we consider the ordinary differential equation defined by:
\begin{equation}
\frac{d{\bm w(t)}}{dt} =  -\frac{\partial{\mathcal{L}(\bm w(t), i)}}{\partial{\bm w(t)}}.
\end{equation}
Define a matrix $\bm H$ where the entry $(i,j)$ is:
\begin{equation}
 \bm{H}_{i,j}(t) =  \left(\frac{\partial{y_{j}^{(L)}(t)}}{\partial{\bm w(t)}}\right) \left(\frac{\partial{ y_{i}^{(L)}(t)}}{\partial \bm w(t)}\right)^T.
\label{eq:theconv}
\end{equation}
Suppose $\bm g_{i} =\frac{\partial y_{i}^{(L)}}{\partial \bm w(t)}$ and $\bm g_{j} =\frac{\partial y_{j}^{(L)}}{\partial \bm w(t)}$.
$\bm{H}_{i,j}(t)$ is the dot-product between two vectors $\bm g_{i}$ and $\bm g_{j}$. Thus, $\bm H$ can be regarded as a Gram matrix. 

Our first step is to calculate the dynamics of each prediction:
\begin{equation}
    \begin{split}
        \frac{d}{dt} y_{j}^{(L)}(t) &=  \langle \frac{\partial{ y_{j}^{(L)}(t)}}{\partial{\bm w(t)}}, \frac{d{\bm w(t)}}{dt} \rangle  \\
  &=   \left(\frac{\partial{y_{j}^{(L)}(t)}}{\partial{\bm w(t)}}\right) \left(\frac{\partial{ y_{i}^{(L)}(t)}}{\partial \bm w(t)}\right)^T  (y_{i}^{*}-  y_{i}^{(L)} ),
    \end{split}
    \label{eq:dyna}
\end{equation}

Given $\bm H$, Eq.~\ref{eq:dyna} can be written as:
\begin{equation}
  \frac{d}{dt} y_{j}^{(L)}(t)  = \bm {H}_{i,j}(t)(y_{i}^* -  y_{i}^{(L)}).
\end{equation}
$\bm{H}_{i,j}(t)$ is the dot-product between two vectors $\bm g_{i}$ and $\bm g_{j}$. We write the dynamic of predictions in a compact way:
\begin{equation}
\frac{d}{dt} \bm y^{(L)}(t) =  \bm H(t)(\bm y^{*} -  \bm y^{(L)}(t)).
\label{eq:grad}
\end{equation}






According to matrix $\bm H$  and Eq.~\ref{eq:grad}, we have the following proposition.

\begin{prop}
Suppose $\bm H(t)$ is the Gram matrix of gradients.  $\forall  t > 0$, the following  inequation holds:
\begin{equation}
\small
  \|( \bm y^{*} - \bm y^{(L)}(t))\|^2_2 
    \leq exp(-\lambda_{min}(\bm H(t))t)\|( \bm y^{*} - \bm y^{(L)}(0))\|^2_2.
\end{equation}
\end{prop}
Refer to Appendix D for detailed proofs. We can see that the upper bound of losses is decided by $\lambda_{min}(\bm H(t))$ and larger  $\lambda_{min}(\bm H(t))$ is expected for lower training losses. 
Furthermore, since $\bm H(t)$ is symmetrical, the F-norm of $\bm H(t)$ bounds $\lambda_{min}(\bm H(t))$ by
\begin{equation}
\lambda_{min}(\bm H(t)) \leq \sqrt{\sum_{i}{|\lambda_i|^2}} = \|\bm H(t)\|_F,
\end{equation}
where $\lambda_i$ is the eigenvalue of $\bm H(t)$. 

As we can see,  the F-norm of GM bounds the convergence rate. 
Each entry in $\bm H(t)$ is the dot-product between two gradient vectors, which is decided by two parts, gradient values and gradient correlations. Intuitively, GM is the health metric of gradients. From the viewpoint of geometric, gradient values decide the step size in optimization, and gradient correlations evaluate gradient directions' randomness. 
Extremely small gradient values will stop training and decrease the convergence rate. Extremely small gradient correlations mean ``random'' gradient directions, 
which bring conflicted parameter updates and increase the training cost. Higher F-Norm values are expected for a higher convergence rate. To conclude, the theoretical results demonstrate that the gram matrix of gradients is an crucial and comprehensive measurement to evaluate the quality of architectures. 

\section{KNAS: Support the Hypothesis}
To support our hypothesis, we propose to leverage GM to evaluate randomly-initialized networks. In implementation, we use the mean of GM, short for MGM, as the final feature and calculate the correlation score between MGM and key optimization features. 

\newtheorem{definition}{Definition}[section]
\begin{definition}[Gradient Kernel]
Suppose $\bm H$ is the Gram matrix of gradients where each item $(i,j)$ represents the dot-product of two gradient vectors. Gradient kernel $\bm  g$ is the mean of all elements in  $\bm H$:
\begin{equation}
 \bm g = \frac{1}{n^2} \sum_{i=1}^n\sum_{j=1}^n\left(\frac{\partial{y_{j}^{(L)}(t)}}{\partial{\bm w(t)}}\right) \left(\frac{\partial{ y_{i}^{(L)}(t)}}{\partial \bm w(t)}\right)^T.
\label{eq:dp}
\end{equation}
\end{definition}


\begin{figure}[t]
\centering
\includegraphics[width=0.7\linewidth]{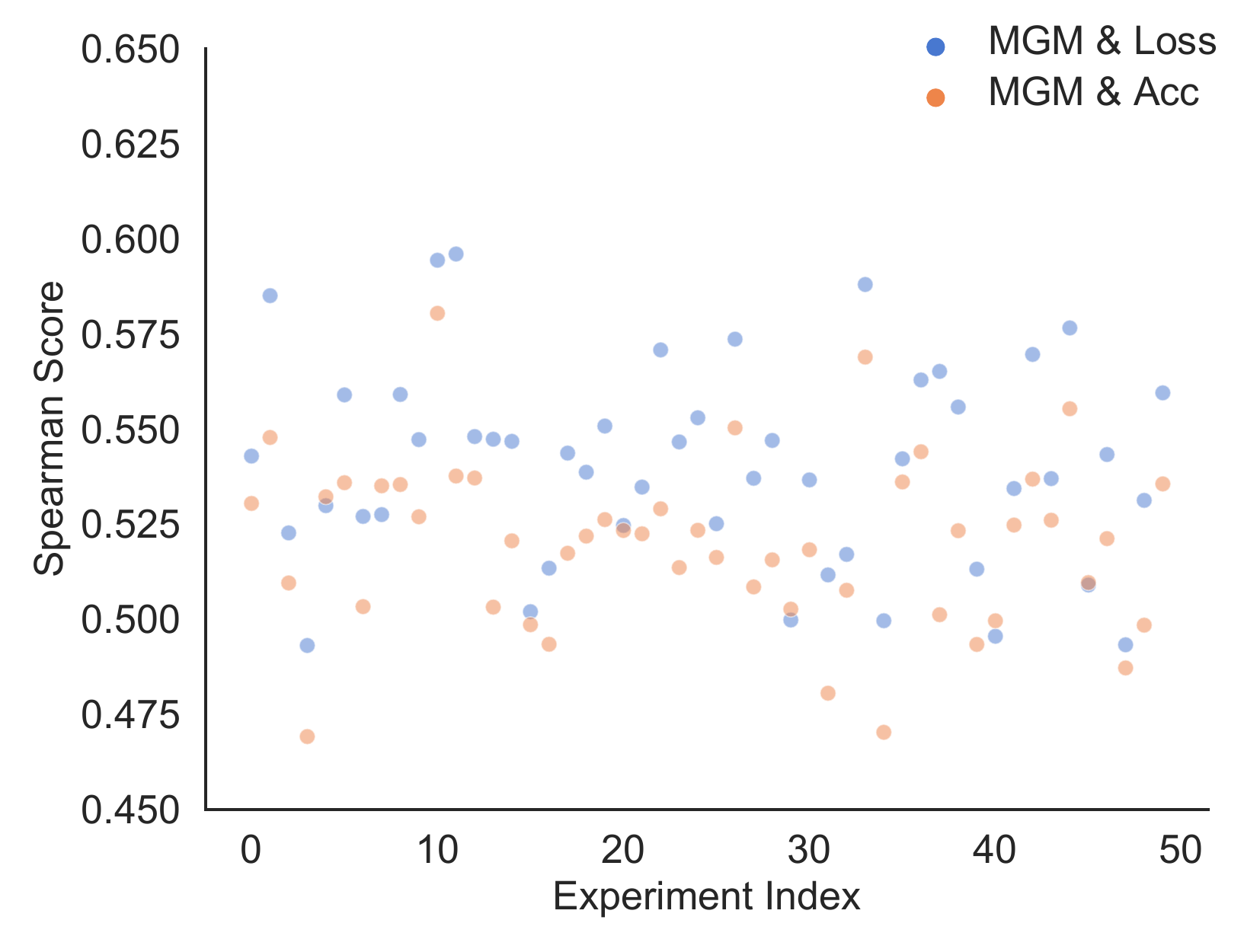}
\caption{The relation between MGM and key optimization results, including training loss and validation accuracy. ``Loss'' represents negative training loss, and ``Acc'' represents validation accuracy. We conduct 50 experiments, each sampling 200 architectures. As we can see, MGM has good positive correlations with training loss and validation accuracy, with around 0.53 and 0.55 Spearman coefficients ($p \ll 0.01 $). }
\label{fig:verify}
\end{figure}

In implementation, since the length of the whole gradient vectors is too long, we approximate Eq.~\ref{eq:dp} as:
\begin{equation}
 \bm g = \frac{1}{Mn^2}\sum_{m=1}^{M} \sum_{i=1}^n\sum_{j=1}^n\left(\frac{\partial{y_{j}^{(L)}(t)}}{\partial{\bm \hat{\bm w}^{m}(t)}}\right) \left(\frac{\partial{ y_{i}^{(L)}(t)}}{\partial \bm  \hat{\bm w}^{m}(t)}\right)^T.
\label{eq:dpimplementation}
\end{equation}
where $K$ is the number of layers. We take each layer as a basic unit and compute the mean of the Gram matrix for each layer. $\hat{\bm w}^{m}$ is the sampled parameters from the $m$-th layer where the length of $\hat{\bm w}^{m}$ is set to $50$ in implementation.

To support the hypothesis, we sample 200 architectures from a NAS benchmark, NAS-Bench-201, and show the relation between kernel values and optimization results on CIFAR100. Figure~\ref{fig:verify} shows that MGM has good correlations with training loss and validation accuracy, which supports our hypothesis.

We then propose a lightweight kernel-based NAS solution, called KNAS. The details are shown in Algorithm~\ref{alg:search}. First, we generate $s$ possible architectures as the search space $\mathcal{S}$. We do not fix the generating policy to keep the flexibility on diverse tasks and scenarios in this work. For each architecture, we calculate MGM based on Eq.~\ref{eq:dpimplementation} and select $k$ architectures with the highest scores as candidates, which are then trained from scratch to get their results on a validation set. To reduce evaluation costs, we adopt validation accuracy at 20 epochs for simplification. 
To be specific, we use the following algorithm to estimate Eq.~\ref{eq:dpimplementation}:
For each parameter, we split its $n$ gradients from $n$ instances/batches into two vectors where each has $n/2$ items. Then, we calculate the dot-product between these two vectors. Finally, we sample $m$ parameter from each parameter matrix and calculate the average dot-product as the final MGM score.

MGM evaluation does not need any training steps. Compared to other NAS approaches that train hundreds of architectures, the new approach can largely reduce search costs. 

\begin{algorithm}[h]
\caption{KNAS Algorithm}
\begin{algorithmic}
\REQUIRE {Search space $\mathcal{S}$, training set $\mathcal{D}_t$, validation set $\mathcal{D}_v$}
 \STATE {Initialize: max\_iteration  = M}
  \STATE {Initialize candidate set $\mathcal{C}$  = []}
 \FOR{search\_iteration in 1, 2, ..., max\_iteration}
  \STATE{Randomly sample an architecture $s$ from  $\mathcal{S}$}
  \STATE{Compute MGM based on Eq.~\ref{eq:dp}}
  \STATE{$\mathcal{C}$.append($s$, MGM)}
  \STATE{update $\mathcal{C}$ to ensure only top-k architectures are kept }
 \ENDFOR
 \STATE{$A^*$ = top\_1($\mathcal{C}$,$\mathcal{D}_t$,$\mathcal{D}_v$)} \# Choose the best architecture based on validation results at 20 epochs. 
 \STATE{return $A^*$}
 \end{algorithmic}
 \label{alg:search}
\end{algorithm}

\section{Experiments}

In this section, we evaluate the proposed approach on NAS-Bench-201~\citep{DBLP:conf/iclr/Dong020}, which provides a fair setting for evaluating different NAS techniques. 

\subsection{Datasets}
\label{sec:dataset}

NAS-Bench-201~\citep{DBLP:conf/iclr/Dong020} is a
benchmark dataset for NAS algorithms, constructed on image classification tasks, including CIFAR10, CIFAR100, and ImageNet-16-120 (ImageNet-16). CIFAR10 and CIFAR100 are two widely used datasets\footnote{https://www.cs.toronto.edu/~kriz/cifar.html}. CIFAR-10 consists of 60,000 32x32 color images in 10 classes, with 6,000 images per class. There are 50,000 training images and 10,000 test images. CIFAR100 has 100 classes containing 600 images each. There are 500 training images and 100 testing images per class. ImageNet-16 is provided by~\citet{DBLP:journals/corr/ChrabaszczLH17}. NAS-Bench-201 chooses four nodes and five representative operation candidates
for the operation set, which generates 15,625 cells/architectures as search space. Each architecture contains full training logs, validation accuracy, and test accuracy on CIFAR10, CIFAR100, and ImageNet-16. In summary, NAS-Bench-201 enables researchers to easily re-implement previous approaches via providing all architecture evaluation results.
However, since these results are unavailable in real-world tasks, we take the evaluation time into account when computing the search time for a fair comparison.

\begin{table*}[t]
\caption{Results on CIFAR10, CIFAR100, and ImageNet-16. ``Time'' means the total search time. The proposed approach KNAS achieves competitive results with the lowest costs. All search steps in KNAS can be finished within a few hours. Results of NASWOT and TE-NAS come from their original papers.  During fine-grained search, KNAS adopts validation accuracy (or training loss) at the 20-th epoch as evaluation criterion. 
 }
\label{tab:nas}
    \centering
     \footnotesize
    \begin{tabular}{c|c|crr|crr|crr}
    \toprule
      \multirow{2}{*}{Type} &  \multirow{2}{*}{Model} & \multicolumn{3}{c|}{CIFAR10} & \multicolumn{3}{c|}{CIFAR100} & \multicolumn{3}{c}{ImageNet-16}  \\
       \cline{3-11}
       & & Acc & Time(s) & Speed-up & Acc & Time(s) & Speed-up & Acc & Time(s) & Speed-up   \\
      \midrule
w/o Search & ResNet&  \textbf{93.97}&N/A  & N/A  & 70.86&N/A  & N/A   & 43.63&N/A  & N/A  \\ 
\midrule
\multirow{5}{*}{ Search}&RS   & 93.63& 216K &1.0x   & 71.28&460K & 1.0x& 44.88& 1M & 1.0x\\
&RL   & 92.83&216K& 1.0x   & 70.71&460K& 1.0x  & 44.10&1M& 1.0x \\
&REA   & 93.72&216K& 1.0x  & \textbf{72.12}& 460K& 1.0x &  45.01& 1M & 1.0x \\
&BOHB & 93.49&216K& 1.0x  & 70.84&460K& 1.0x  & 44.33&1M & 1.0x\\
&RSPS & 91.67&10K&21.6x  & 57.99&46K &21.6x& 36.87& 104K& 9.6x\\
\midrule
\multirow{2}{*} {Gradient} &
GDAS  & 93.36&22K&12.0x   & 67.60& 39K&11.7x & 37.97&130K &7.7x  \\
& DARTS &  88.32 & 23K &9.4x  & 67.34& 80K&5.8x & 33.04& 110K& 9.1x  \\
\midrule

\multirow{2}{*} {Training-free}  & NASWOT & 92.96 & 2.2K & 100x & 70.03 & 4.6K & 100x & 44.43 &  10K & 100x\\ 
& TE-NAS & 93.90 & 2.2K & 100x  & 71.24 & 4.6K  & 100x & 42.38 & 10K & 100x \\ 
\midrule

\multirow{2}{*} {MGM}  & \bf KNAS ($k=20$) & 93.38  &4.4K& 50x  & 70.78 & 9.2K & 50x & 44.63 & 20K & 50x \\
 & \bf KNAS ($k=40$) & 93.43 & 8.8K &25x  & 71.05&18.4K &25x & \bf 45.05& 40K & 25x \\
       \bottomrule
    \end{tabular}
    \label{tab:my_label}
\end{table*}

\subsection{Baselines}
\label{sec:baseline}

We compare the proposed approach with the following baselines:
\paragraph{Random Search Algorithm} It includes two baselines, random search (RS)~\citep{DBLP:journals/jmlr/BergstraB12} and random search with parameter sharing (RSPS)~\citep{DBLP:conf/uai/LiT19}. 
Random search is a simple baseline that randomly samples several architectures and selects the network with the best validation accuracy as the final architecture. 
RSPS introduces a two-stage training process. The first is a shared-parameter training stage. The second is a search stage, which chunks pre-trained modules and evaluates all possible architectures on a validation set. The architecture with the highest validation accuracy is selected as the final architecture. 

\paragraph{Reinforcement learning based algorithm} It includes one baseline, RL~\citep{DBLP:journals/ml/Williams92}. RL is a method that introduces an architecture generator to generate well-performing architectures. To train the generator, it uses validation accuracy as reward. 

\paragraph{Evolution-based search algorithm} It includes one baseline: regularized evolution for architecture search (REA)~\citep{DBLP:conf/aaai/RealAHL19}. It first selects and trains several parent architectures and then applies mutation operations on these parent architectures to get child architectures. The architecture with the best validation accuracy is adopted as the final architecture.


\paragraph{Differentiable algorithm} It includes two baselines, DARTS~\citep{DBLP:conf/iclr/LiuSY19} and GDAS~\citep{DBLP:conf/cvpr/DongY19}. DARTS  reformulates the search problem into a continues search problem. It uses one-batch gradients to replace the accuracy on a validation set to guide a model how to search for well-performing architectures. Due to unsteady training, the results provided by the original NAS-Bench-201 paper are very low.  To avoid model collapse, we use different seeds to train the whole framework multiple times and generate multiple candidate architectures, which are then trained from scratch to select the best one. 
Following this work, GDAS updates a sub-graph rather than the whole graph for higher speedups. 

\paragraph{Hyper-parameter optimization algorithm}  It includes one baseline, BOHB~\citep{DBLP:conf/icml/FalknerKH18}. This approach combines the benefits of Bayesian optimization and bandit-based methods. It requires ``train-then-test'' paradigm  to evaluate different architectures. 
We use the code provided by~\citet{DBLP:conf/iclr/Dong020} for baseline implementation. 

\paragraph{Training-free algorithm} It includes two approaches, NASWOT and TE-NAS. NASWOT uses the  output of rectified linear units to evaluate architectures.  TE-NAS uses the spectrum of the neural tangent kernel and the number of linear regions in the input space to evaluate architectures. 

\begin{figure*}[t]
    \centering
    \vspace{-0cm}
    \includegraphics[width=0.7\linewidth]{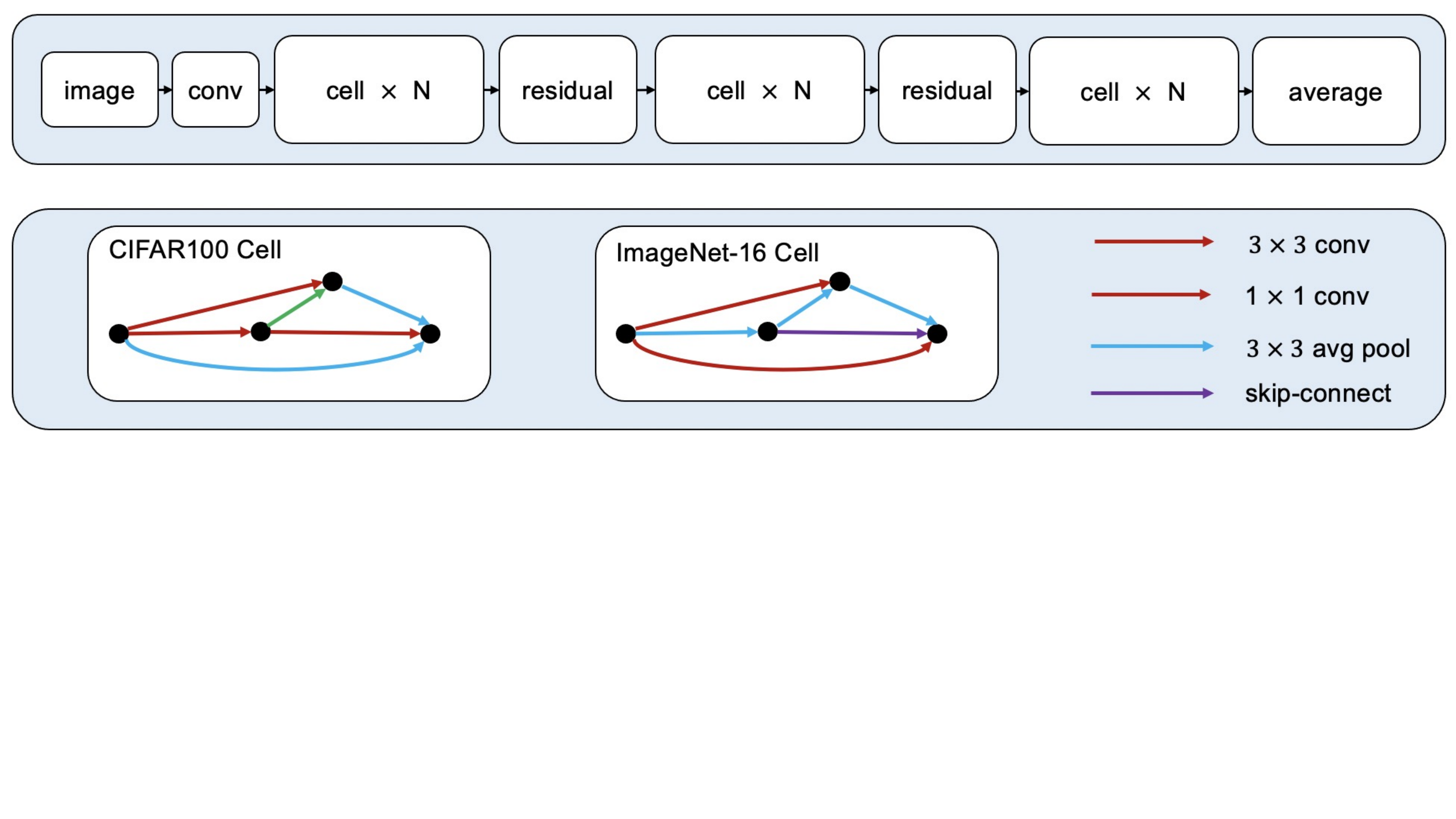}
    \vspace{-3cm}
    \caption{The searched architectures for CIFAR100 and ImageNet-16. They share the same architecture framework (shown in the top block). The bottom block shows the details of the searched cell. }
    \label{fig:structure}
\end{figure*}

\subsection{Experiment Settings}
We randomly sample 100 architectures as the search space in the proposed approach and calculate their MGM scores at initialization.    All baselines are implemented on a single NVIDIA V100 GPU. 
We use the released code provided by~\citet{DBLP:conf/iclr/Dong020} for baseline implementation. 

 Architecture evaluation contains two steps:  network training and evaluating.
For all approaches, we set the time of architecture training plus evaluation to 2,160 seconds, 4,600 seconds, and 10,000 seconds on CIFAR10, CIFAR100, and ImageNet-16, respectively. Here we do not adopt the reported time in~\citet{DBLP:conf/iclr/Dong020} for architecture evaluation because the authors use multiple workers to train a model. In contrast, we adopt a single GPU (Tesla-V100) and a single worker. 
 We observe that architecture evaluation takes almost all search time in search-based approaches (including RS, RL, REA, BOHB, RSPS). The rest operations only take a few seconds, which are not included in the search time for simplicity.  For DARTS and RSPS, we slightly update the search policy for better performance. We run DARTS and RSPS up to $10$ epochs and evaluate the selected architecture after each epoch. 
 

\begin{figure}[t]
\centering
\subfigure[CIFAR100]{
\begin{minipage}[b]{0.40\linewidth}
\includegraphics[width=1\linewidth]{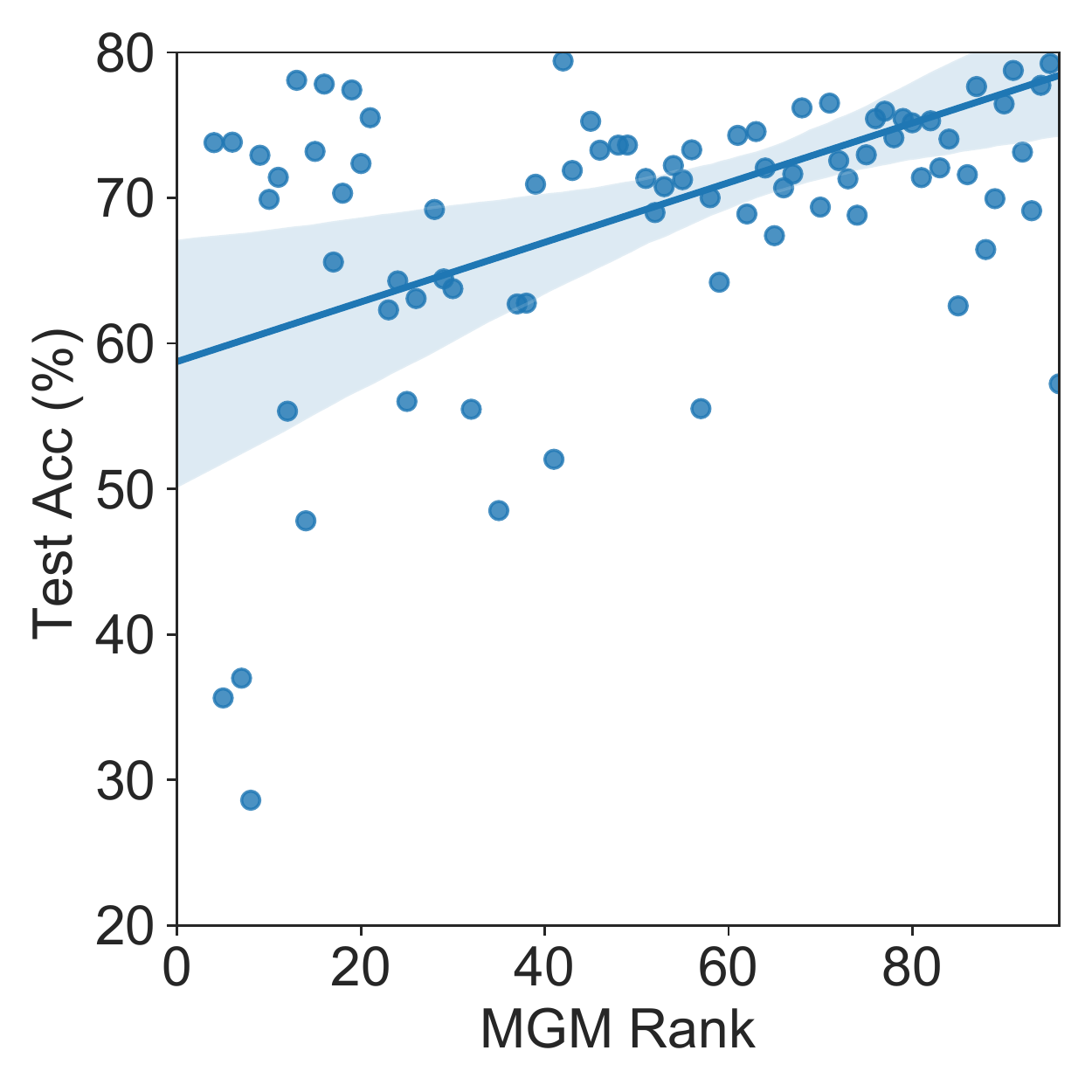}
\end{minipage}
}
\subfigure[ImageNet-16]{
\begin{minipage}[b]{0.40\linewidth}
\includegraphics[width=1\linewidth]{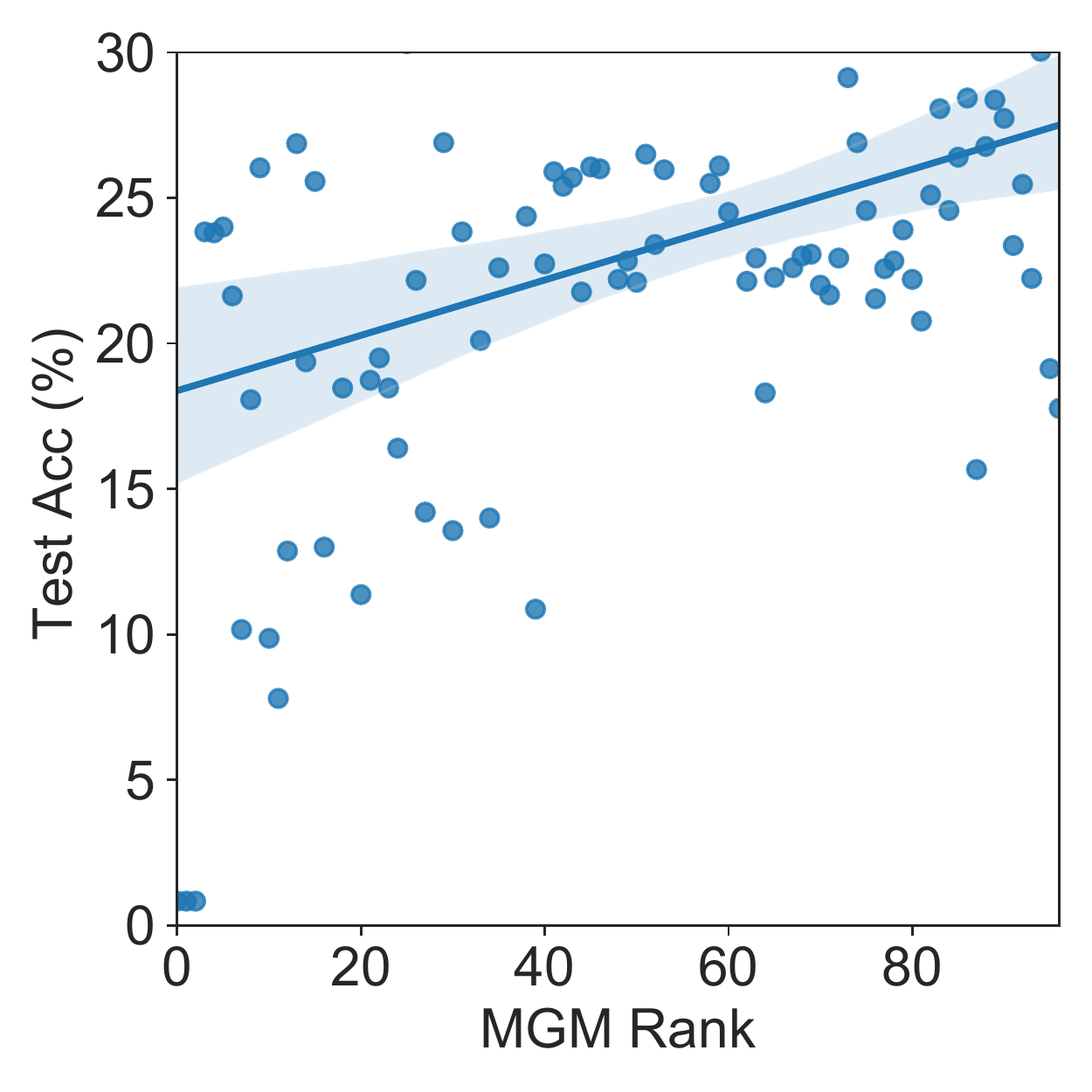}
\end{minipage}
}

\caption{MGM has good correlations with downstream performance on CIFAR100 and ImageNet-16. Smaller rank represents smaller MGM. Y-axis lists test accuracy.  }
\label{fig:dot}
\end{figure}

\section{Results}

Table~\ref{tab:nas} shows the comparison between the proposed approach and strong baselines. RS is a naive baseline, which randomly selects architectures for full-training and evaluation. It is one of the most time-consuming methods and achieves good accuracy improvements with 0.46 and 1.25 gains on CIFAR100 and ImageNet-16. RL adds reinforcement learning into the search process and is able to explore more well-performing architectures. However, due to its unsteady training, RL does not beat RS on all datasets in terms of accuracy. REA contains two search stages: parent search and mutation. The second stage explores new architectures based on the best-performing parent architectures. Roughly speaking, better parent networks have better inductive bias and thus provide student networks with good initialization. This approach achieves the highest performance with 93.72, 72.12, and 45.01 scores on CIFAR10,  CIFAR100, and ImageNet-16. However, despite promising results, these baselines require considerable time for architecture evaluation. Besides, several approaches also achieve much better speed-up results, such as RSPS, GDARTS, and DARTS.
However, the accuracy of architectures searched by these approaches is the main problem. 

Unlike these approaches, the proposed approach achieves competitive results with the lowest costs. Compared to the time-consuming approaches, including RS, RL, REA, and BOHB, KNAS with $k$=40 brings around 25x speedups on CIFAR10, CIFAR100 ImageNet-16. Compared to approaches with lower search costs, including RSPS, GDAS, DARTS, NASWOT, and TE-NAS, the proposed approach achieves large accuracy improvements on Imagenet-16. The searched architecture also outperforms ResNet with 0.19 and 1.42 accuracy improvements on CIFAR100 and ImageNet-16. Figure~\ref{fig:structure} visualizes the architectures searched by KNAS. Furthermore, we also visualize the relation between MGM and downstream performance on CIFAR100 and ImageNet-16. Results are shown in Figure~\ref{fig:dot}.  Each dot represents a single architecture with its MGM value (X-axis) and accuracy (Y-axis). As we can see, the proposed hypothesis also holds on IFAR10 and ImageNet-16. They both have good correlations between MGM and downstream performance.


\begin{figure*}[t]
    \centering
    \includegraphics[width=0.65\linewidth]{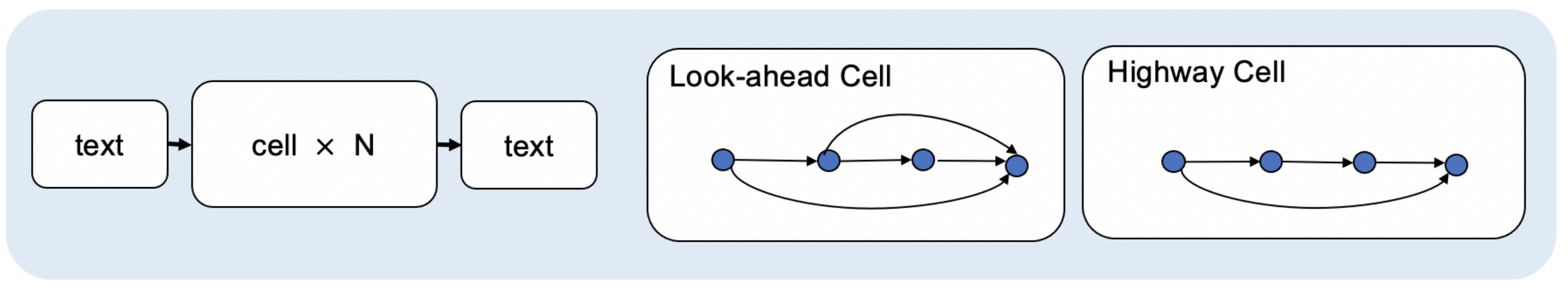}
    \caption{An illustration of look-ahead networks and high-way networks.  They have the same architecture framework where all cells are connected in a chain way. For each look-ahead cell, the last layer takes all inner outputs as the input. Highway cells are similar to look-ahead cells except for the last layer only taking the first layer output and the last layer output as the input.}
    \label{fig:res}
\end{figure*}

\section{Discussion: Generalization on Diverse Tasks}
To avoid the bias on specific factors, like initialization, we conduct more experiments on text classification tasks to verify the generalization ability of KNAS. The settings of text classification are different from that of NAS-Bench-201. For example, NAS-Bench-201 adopts Kaiming initialization, CNN cells, batch normalization, while text classification adopts Xavier initialization, self-attention cells, layer normalization.  Text classification includes two datasets, MRPC evaluating whether two sentences are semantically equivalent and RTE recognizing textual entailment. Two classification datasets are provided by~\citet{DBLP:conf/iclr/WangSMHLB19}. 



\subsection{Datasets and Baselines}

Text classification datasets is provided by GLUE~\citep{DBLP:conf/iclr/WangSMHLB19}, a platform designed for natural language understanding. We use MRPC and RTE in this work. MRPC is a classification dataset for evaluating whether two sentences are semantically equivalent. Each example has two sentences and a human-annotated label. We use accuracy as the evaluation metric. RTE is a textual entailment dataset. It comes from a series of annual textual
entailment challenges. Following~\citet{DBLP:conf/iclr/WangSMHLB19}, we combine the data from RTE1, RTE2, RTE3, and RTE5 and adopt the same data-processing steps.  We adopt RoBERTA-large, a widely-used pre-trained model, as the baseline. The code is provided by HuggingFace\footnote{https://github.com/huggingface/transformers}, a widely used pre-training model project.    All network candidates are initialized with pre-trained parameters. We run Roberta-large on 8 V100 GPUs. For each dataset, we randomly run  Roberta-large $5$ times and report the average results. For MRPC, the batch size is set to $4$, and the learning rate is set to $3e-5$. For RTE, the batch size is set to $4$, and the learning rate is set to $2e-5$. For the rest hyper-parameters, we use the default settings. 


\subsection{Search Space}
\label{sec:searchspace}
We define several skip-connection structures, each with possible candidate architectures, as shown in Figure~\ref{fig:res}. We merge these architectures as a search space. We adopt the same settings for all architectures: 12 encoder layers and 12 decoder layers. These architectures also share the same hype-parameters.  Here are the details of different structures.

\paragraph{Highway networks}  It connects all cells in a chain way. In each cell, there exists a highway connection from the first layer to the last layer.  The number of layers in each cell ranges from $2$ to $11$, and there are $10$ possible architectures.


\paragraph{Look-ahead networks} In each cell, the layer is connected in a chain way except for the last layer. The last layer takes all outputs of previous layers in a cell as inputs. The number of layers in each cell ranges from $2$ to $11$, and there are $10$ possible architectures.


\paragraph{DenseNet} It splits the whole network into different cells~\citep{huang2017densely}. We adopt the original DenseNet structure and omit its implementation details in Figure~\ref{fig:res} for simplification.  The input of the current cell comes from all outputs of previous cells. In each cell, the layer is connected in a chain way.   The number of layers in each cell ranges from $2$ to $11$.

\subsection{Results}
The results are shown in Table~\ref{tab:bias}. The architectures searched by the proposed approach achieve better results on all datasets. Though pre-trained models are widely believed to be  state-of-the-art approaches, KNAS still achieves performance improvements over RoBERTA-large with $1.24$ and $0.24$ accuracy improvements on  MRPC and RTE datasets. We choose top-$2$ and top-$5$ architectures for full training and evaluation for two classification datasets.  It proves that the proposed approach generalizes well on various datasets.

\begin{table}[t]
\centering
\caption{ Results on text classification tasks. ``Time'' refers to the search time.    }
 \footnotesize
 \begin{tabular}{c|cc|cc}
  \toprule  
 
\multirow{2}{*}{Models}  & \multicolumn{2}{c|}{ MRPC}  & \multicolumn{2}{c}{RTE} \\
& Acc & Time(s) & Acc & Time(s) \\
\midrule
  
  RoBERTA-large &  92.08& N/A & 83.51& N/A  \\
  
  KNAS & \textbf{93.32}& 0.4K & \textbf{83.75} &2K    \\
   \bottomrule

  \hline 
  \end{tabular}
  
\label{tab:bias}
\end{table}

\section{Conclusion}
\label{sec:conclusion}
In this work, we aim to explore a green NAS solution by getting rid of downstream training from architecture evaluation. We propose a hypothesis that gradients can be used to evaluate randomly-initialized networks. To support the hypothesis, we conduct a theoretical analysis and find an appropriate feature MGM. According to this feature, we develop a simple yet efficient architecture search approach KNAS. It achieves large speedups with competitive accuracies on a NAS benchmark dataset. 
Furthermore, KNAS generalizes well and can search for better architectures on text classification tasks.

\section*{Acknowledgement}
We thank the anonymous reviewers for their constructive suggestions and comments. This work is partly supported by Beijing Academy of Artificial Intelligence (BAAI).

\bibliographystyle{icml2021}
\bibliography{icml2021}

 \clearpage

\section*{Appendix D: Proofs for Proposition 4.1}
\label{pr:pro}

Recall we can write the dynamics of predictions as
\begin{equation}
\small
\frac{d}{dt} \bm y^{(L)}(t) =  \bm H(t)( \bm y^{*} -  \bm y^{(L)}(t)).
\end{equation}


By spectral theorem, we can write $\bm H(t) =\bm P\bm D\bm P^T$ where $\bm P$ is orthogonal and $\bm D>0$ is diagonal with each entry being eigenvalue of $\bm H(t)$. We observe that
\begin{equation}
\begin{split}
\small
   &\min_{\|\bm x\|=1} \bm x^T \bm H(t) \bm x  \\
   & =  \min_{\|\bm x\|=1} (\bm P^T\bm x)^T \bm H(t) (\bm P^T \bm x) \\ & = \min_{\|\bm y\|=1} \bm y \bm D \bm y^T \\ & = \lambda_{min}(\bm H(t))
\end{split}
\label{eq:eqv}
\end{equation}
Define
\begin{equation}
\small
    Q = \|\bm y^{*} - \bm y^{(L)}(t)\|^2_2.
    \label{eq:q}
\end{equation}
Given $Q$, we can calculate the loss function dynamic as:
\begin{equation}
\begin{split}
\small
&\frac{d}{dt}\|(\bm y^{*} - \bm y^{(L)}(t) )\|^2_2 \\&  = -(\bm{y}^{*} - \bm{y}^{(L)}(t))\frac{d}{dt}\bm y^{(L)}(t)\\
& = -(\bm{y}^{*} - \bm{y}^{(L)}(t))\bm H(t)( \bm y^{*} -  \bm y^{(L)}(t)) \\
 & = -Q\frac{(\bm y^{*} - \bm y^{(L)}(t))^{T}}{\|\bm y^{*} - \bm y^{(L)}(t)\|}\bm H(t)\frac{\bm y^{*} - \bm y^{(L)}(t)}{\| \bm y^{*} - \bm y^{(L)}(t)\|}.
\end{split}
\end{equation}
Based on Eq.~\ref{eq:eqv} and Eq.~\ref{eq:q}, the following inequation holds:
\begin{equation}
\begin{split}
\small
     &\frac{d}{dt}\|( \bm y^{*} - \bm y^{(L)}(t))\|^2_2  \\
     &= - Q\frac{(\bm y^{*} - \bm y^{(L)}(t))^{T}}{\|\bm y^{*} - \bm y^{(L)}(t)\|}\bm H(t)\frac{\bm y_i^{*} - \bm y^{(L)}(t)}{\| \bm y^{*} - \bm y^{(L)}(t)\|} \\
    &\leq -\lambda_{min}(\bm H(t))\|(\bm y^{*} - \bm y^{(L)}(t))\|^2_2.
\end{split}
\label{eq:se1}
\end{equation}

Under Assumption 5.1, $\lambda_{min}(\bm H(t)) > 0$. Therefore, $\frac{d}{dt}\|(\bm y^{*} - \bm y^{(L)}(t))\|^2_2 \leq 0$ holds.
  
Here, we use $\lambda_{min}$, short for $\lambda_{min}(\bm H(t))$. The expectation of gradients of $exp(\lambda_{min} t)\|(\bm y^{*} - \bm y^{(L)})\|^2_2$ is
\begin{equation}
    \begin{split}
    \small
    &\frac{d}{dt}exp(-\lambda_{min} t)\|\bm y^{*} - \bm y^{(L)}\|^2_2 \\
    &= -\lambda_{min}exp(-\lambda_{min} t)\|\bm y^{*} - \bm y^{(L)}\|^2_2 \\ & + exp(\lambda_{min} t)\frac{d}{dt}\|\bm y^{*} - \bm y^{(L)}(t)\|^2_2 \\
    &\leq-\lambda_{min}exp(-\lambda_{min} t)\|\bm y^{*} - \bm y^{(L)}\|^2_2 \\ & - exp(-\lambda_{min} t)\lambda_{min}\|\bm y^{*} - \bm y^{(L)}(t)\|^2_2 \\
        &\leq 0.
    \end{split}
\end{equation}
Since the upper bound of the gradients is negative, $\forall  t > 0$, the following inequation holds:
\begin{equation}
\begin{split}
\small
    exp(\lambda_{min}t)\|\bm y^{*} - \bm y^{(L)}(t
    )\|^2_2 &\leq exp(0)\| \bm y^{*} - \bm y^{(L)}(0)\|^2_2 \\
    &\leq \| \bm y^{*} - \bm y^{(L)}(0)\|^2_2,
    \end{split}
\end{equation}
and thus,
\begin{equation}
\begin{split}
\small
    \|\bm y^{*} - \bm y^{(L)}(t)\|^2_2 &\leq \frac{\| \bm y^{*} - \bm y^{(L)}(0)\|^2_2}{exp(\lambda_{min}t)}  \\
    &\leq exp(-\lambda_{min}t)\|\bm y^{*} - \bm y^{(L)}(0)\|^2_2.
    \end{split}
\end{equation}
$\hfill\qed$

\end{document}